\documentclass[runningheads]{llncs}
\usepackage{booktabs}
\usepackage{subfigure}
\usepackage{xcolor}
\usepackage[T1]{fontenc}
\usepackage{graphicx}
\usepackage{url}

\usepackage{hyperref}
\hypersetup{
    colorlinks=true,
    urlcolor=blue,
    linkcolor=blue,
    citecolor=blue
}

\begin{document}
\title{Democratizing the medieval English legal tradition}
\author{Michael Zhang\inst{1}\orcidID{0000-0002-0659-1783} \and
Elise Wang\inst{2}\orcidID{0009-0009-0438-6565} \and
Charlotte Whatley\inst{3}\orcidID{0009-0000-6181-927X} \and
Seth Strickland\inst{4}\orcidID{0000-0002-7039-9163} \and
Dylan Bannon\inst{5}\orcidID{0009-0003-6791-8166}
}
\authorrunning{M. Zhang et al.}
\institute{University of Chicago, Chicago IL 60637
\email{mzzhang2014@gmail.com} \\
\url{https://astro.uchicago.edu/~mz/} \and
California State University, Fullerton CA 92831 \and
University of Wisconsin-Madison, Madison WI 53706 \and
Carnegie Mellon University, Pittsburgh PA 15213 \and
Unaffiliated
}

\maketitle              
\begin{abstract}
The record of the beginning of the most widespread legal system in the world is contained in millions of pages of handwritten text.  Most of the records of the first centuries of the Anglo-American legal system are hand-written in a highly abbreviated form of medieval Latin which only a few dozen scholars in the world are trained to read.  In this interdisciplinary project, we construct a dataset of 4029 lines of text across 193 medieval criminal and civil cases.  We then use the dataset to train an open-source end-to-end pipeline for transcribing these manuscripts.  We first train standard neural network architectures for line segmentation and handwriting recognition (R-Blla and CNN+LSTM with CTC decoding, respectively) and show that they can already achieve 79\% word accuracy, despite the relatively small training set and the challenge of expanding abbreviations.  We then demonstrate that simple post-processing significantly boosts accuracy: adding an n-gram language model to the CTC decoder improves word accuracy to 82\%, while asking Gemini Pro 3 to correct mistakes boosts accuracy to 88\%.  Finally, we compare the CNN+LSTM architecture with TrOCR, a transformer-based OCR architecture, demonstrating that TrOCR shows comparable word accuracy but worse character accuracy due to its over-willingness to guess, making it harder for humans to infer the correct reading.  We incorporated our pipeline into a web portal (glyphmachina.com), opening up the English legal tradition to legal scholars, medievalists, and students.
\keywords{Handwritten Text Recognition  \and Latin}
\end{abstract}
\section{Introduction}
The English common law emerged as a unified legal system over the course of the twelfth and thirteenth centuries.  Common law functions through the accumulation of cases; in the most basic sense, each case provides guidance for how similar, future cases ought to be decided. As a result, records are of paramount importance to the system, and were meticulously preserved, both at the king's court and in local archives. Over centuries, this body of law formed the basis for the common law system that persists to this day and has left a lasting legacy on every continent in the world.  It is the basis of the legal system not just in the United Kingdom, but also in many of its former colonies, including the United States, India, and Tanzania.  It is estimated that 35\% of the world's population live under either a common law system or a mixed system with common law as one of the main components.\footnote{\url{https://www.juriglobe.ca/eng/syst-demo/tableau-dcivil-claw.php}}

Despite the immense impact of this tradition, very few historians are able to directly study the medieval cases on which it is based.  In recent years, this has not been due entirely to lack of access.  In fact, most of the extant legal record from before 1500 (preserved on parchment in a handwritten script) has been digitized and made available online at the Anglo-American Legal Tradition (AALT) website (\url{http://aalt.law.uh.edu/}). To read these sources, however, one must be trained in medieval legal Latin, in the idiosyncratic system of abbreviation that these records use, and in deciphering the handwriting of scribes who were writing more for speed than long-term legibility. Furthermore, mere access is not enough to use the archives -- scholars need an instinct for what is important, decades of experience in the currently available records, and support or budget to comb through these millions of important pages. Even scholars who have such access rarely use the online archives alone, since they operate better as a digital reference database than a browse-able archive. Thus, though eleven million pages of the legal manuscript records at the National Archives in Kew, London are available online, only a few dozen scholars in the world can make use of them. Around twenty of these rolls are transcribed, translated, and published, but this is a tiny minority of the extant corpus. This means that these records are largely inaccessible to students not yet proficient in paleography, legal scholars who specialize in other periods, and even medievalists who do not regularly work with these archives.

These manuscripts pose a number of unique challenges.  Some are written in Anglo-Norman, the language of prestige in post-Conquest England, but the majority are written in Latin.  Due to the scribes' sometimes limited Latin proficiency, the syntax and spelling are not always correct.  The script is highly abbreviated, so that around half of the words have one or multiple letters omitted.  As is typical of medieval scribal productions, some abbreviations are predictable but most are not standardized.  Sometimes abbreviations are indicated with a bar or a loop; sometimes they are implicit.  Even when words are not abbreviated, letter forms often merge and are often difficult or impossible to distinguish.  For example, ``i'', ``n'', and ``m'' are usually indicated with 1, 2, and 3 short vertical lines (``minims'') respectively, making it impossible to correctly read a succession of minims without combining the linguistic and syntactic contexts.  To top off the challenges, even within a script family, different scribes have different handwriting styles, abbreviation choices, and letter forms, requiring adaptive recontextualization even for expert users.  These difficulties mean that even a trained paleographer can require around five minutes to decipher a single line of text, making archival searches painstaking and time consuming.

In this work, we use artificial intelligence to enable an end-to-end pipeline to transcribes medieval English legal manuscripts quickly and accurately.  We also release the dataset we used to train the AI in PageXML format.  To our knowledge, it is the only publicly available dataset of medieval English legal text amenable to training neural networks.

\section{Related Works}
Handwriting text recognition (HTR) has long been a field of active research.  We highlight some milestones here, but refer the reader to \cite{garrido-munoz_2025} for a more detailed survey.  The first commercial products incorporating handwriting recognition were introduced in the 1980s.  In the 1990s, the harder problem of recognizing handwritten text from an image was usually tackled using a combination of character-level segmentation and handcrafted features (e.g. \cite{kim_1997}).  The advent of powerful programmable GPUs ushered in the deep learning era, enabling the training of neural networks to transcribe entire lines of text with impressive accuracy without any handcrafted features.  In 2015, \cite{shi_2017} first proposed the deep learning architecture that would become standard: a convolutional neural network feeding into a bidirectional long short-term memory (LSTM) recurrent neural network, all trained with CTC loss.  In 2017, the attention-based transformer architecture \cite{vaswani_2017} ushered in a new paradigm and revolutionized deep learning.  The first work to utilize the transformer architecture for HTR was \cite{kang_2022}, who obtained state-of-the-art error rates by using a ResNet CNN as a visual feature extractor and a series of multi-head attention layers in both the encoder and decoder.  \cite{diaz_2021} obtained even better performance with a CNN backbone coupled with a self-attention encoder and a CTC decoder.  In 2021, Microsoft released TrOCR \cite{li_2023}, which dispenses with CNNs entirely and uses the transformer architecture for both image processing and text generation.  In 2023, \cite{fujitake_2024} introduced DTrOCR, which achieves substantially lower error rates than all previous models by removing the encoder in TrOCR and using only the decoder.  The input image is broken up into 16x16 patches, which are converted into vectors with a patch embedding before being fed to a decoder initialized with GPT-2 weights.  The model is pre-trained with two billion synthetic line images in 5427 handwritten fonts, and then fine-tuned on real data.  The pre-trained DTrOCR model is, however, not public.

In parallel to these algorithmic improvements, researchers have developed HTR platforms such as Transkribus \cite{kahle_2017}, kraken,\footnote{\url{https://kraken.re/main/index.html}} and PyLaia \cite{puigcerver_2018}.  These platforms allow users unfamiliar with ML-specific tools such as PyTorch or TensorFlow to either train a neural network on their own data, or use a pretrained network from the literature.  Transkribus has seen especially wide adoption across many disciplines \cite{nockels_2022}.

A few works have focused specifically on handwriting recognition with medieval manuscripts.  \cite{aguilar_2023} trained a CNN + LSTM architecture with CTC loss on 120,000 lines of French and Latin, obtaining a final word error rate of 16\% for the Textualis script and 28\% for the Cursiva script.  \cite{clerice_2024} released CATMuS, a dataset of 160,000 text lines in 10 different languages with Latin scripts, spanning the 8th to 16th centuries.  They also released kraken and PyLaia models trained on their dataset.  \cite{aguilar_2025} constructed an open-source corpus of medieval and early modern manuscripts (TRIDIS) spanning the 12th to 16th centuries and trained two HTR models, TrOCR and MiniCPM-Llama3-V.  They achieve a WER of 21--25\% and a CER of 9--11\% with TrOCR. 

Recently, \cite{coll-ardanuy_2026} evaluated state-of-the-art off-the-shelf models on a new dataset, including models trained on the CATMuS and TRIDIS datasets.  They find that these models are still far from providing a generalizable transcription solution, but provide a good foundation for fine-tuning to task-specific datasets.


\section{Dataset}
For our dataset, we transcribed 4029 lines (3602 for training, 427 for testing) from cases drawn from the rolls digitized on the AALT archive. Each case was handwritten in abbreviated Latin in a script called \textit{cursiva anglicana} or chancery cursive on a “membrane” of parchment. These membranes were then sewn together at the top (like a legal pad) into “roll,” which were organized by the court they came from, the type of case, and the term in which they occurred. In general, we sought cases that would represent the archive as it is most useful to a researcher. This meant that we focused on the period from Edward I to Henry VI (1272-1461), which uses a consistent script and relatively consistent forms. These records are far more numerous and far less organized than those of the twelfth and early thirteenth centuries, and as a result only a small percentage of them have been transcribed.

Among Latin datasets, our dataset offers an unusual corpus that, despite its size, is a better representation of “everyday” documentary Latin than other existing datasets. Latin was the documentary language of the medieval European world, which meant that almost all legal, financial, trade, and property documents were written in Latin. Most datasets (and collections of Latin texts, like the \textit{Corpus Corporum}) collect texts that have been important to humanities researchers — literature, philosophy, and political treatises — but that are not terribly representative of everyday life. 

Among manuscript HTR datasets, ours is again unique, in part due to the formulaic nature of these documents, the standardization of the vocabulary, and the large size of the corpus. Scripts change over time, sometimes rapidly, but in this case, because these records formed the basis of a legal system that constantly referenced its own history, the script stayed relatively stable for at least 200 years. The specialized vocabulary and abbreviations in these records are legion but also relatively stable, and as a result, the 4000 lines we extracted are representative enough of a corpus of at least 220 million (11 million images, each with around 20-35 lines on them).

\subsection{Data preparation}
Rather than working at the roll or membrane level, we chose individual cases, because most researchers read the rolls in that way, and because it allowed us to sample the dataset much more widely. On average, we have not included more than three cases from the same roll. In our selection, we chose cases that would give us two types of variety: case type (varying the crime, severity, etc.) and “hand” (the individual handwriting style). In search of a representative sample, we drew from three types of rolls: the King’s Bench Rolls (KB), the Common Plea Rolls (CP), and the eyre rolls (JUST1).  We also included the 51 example lines that AALT uses to showcase Latin palaeography.\footnote{\url{http://aalt.law.uh.edu/Pal1.html}, \url{http://aalt.law.uh.edu/Pal2.html}, \url{http://aalt.law.uh.edu/Pal3.html}}

\begin{table}
\centering
\caption{Number of cases, lines, and words from each type of roll in the dataset.}
\setlength{\tabcolsep}{8pt}
\begin{tabular}{@{} l c c c @{}}
\hline
  Roll Type & Cases & Lines & Words\\
\hline
King's Bench (KB) & 54 & 916 & 15,495\\
Common Pleas (CP) & 39 & 887 & 14,746\\
Justices Itinerant (JUST1) & 91 & 2175 & 38,619\\
Palaeography Assistance & 9 & 51 & 892\\
\hline
\end{tabular}
\label{table:performance_comparison}
\end{table}

\begin{figure}
    \centering
    \subfigure{\includegraphics[width=1\linewidth]{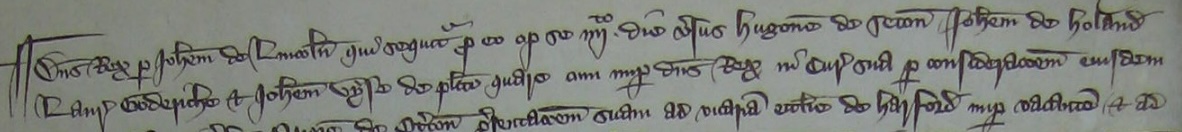}}
    \subfigure{\includegraphics[width=1\linewidth]{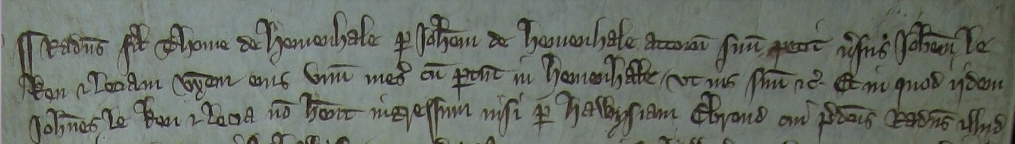}}
    \subfigure{\includegraphics[width=1\linewidth]{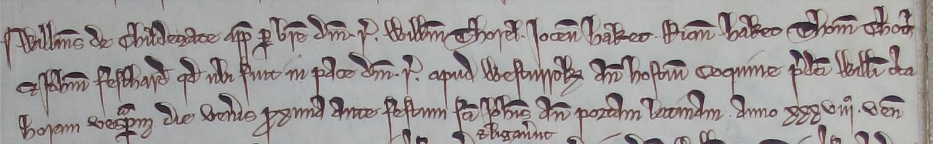}}
    \caption{Samples from our dataset: KB27 349m21, CP40 169m62, JUST1 235m13.}
    \label{fig:dataset_samples}
\end{figure}

The King’s Bench was the most senior criminal court of England. We drew primarily from KB27, known as the \textit{Coram Rege} rolls, which provide a mix of subject, place, and stage of process. The Common Pleas Rolls were also crown business, but include a wider variety of cases. The scribes of the KB and CP rolls were well-trained and tend to be regular, use standard abbreviations, and follow a familiar structure (top panel of Figure \ref{fig:dataset_samples}). Because the KB and CP rolls are especially important to the formation of the common law, about half of our data is drawn from these rolls. The JUST1 rolls were the records of the justices in eyre, or the court that the crown sent across the island to try serious cases at each county. These records are more idiosyncratic in almost all ways: in their organization and indexing, in their vocabulary, and in the types of cases they recorded. JUST1 hands tend to be messier, more inconsistent in abbreviation, and sometimes follow English word order because the scribe translated on the fly into Latin (see bottom panel). Most scholars accept Ralph Pugh’s estimate that criminal cases in eyre took on average fifteen minutes \cite{pugh_1983}, during which time the scribe would also complete his record. All of these features make them harder to transcribe for the average paleographer, so we included a large number of them in order to train the model on this more difficult type with the aim of making the type accessible.

\subsection{Constraints in data preparation}
A major constraint on the dataset has been human labor. EW and CW performed the transcriptions by hand and cross-checked each other's work, but this process is extremely time-consuming. Both estimate that a case of around fifteen lines took at minimum an hour to transcribe correctly, and another fifteen minutes to check for correctness. SS is trained in paleography, but primarily in medieval English and non-legal Latin. A case might take him hours; unfamiliarity with legal abbreviations in these cases would make repeated consultation with EW or CW necessary, thus limiting the efficacy of a transcriber ostensibly trained in a similar skillset. We point this out to indicate just how specialized working knowledge of these scripts are. Predictably, the KB and CP rolls sometimes went faster, and the JUST1 rolls much slower.

Drawing from the AALT archive has been fruitful but has also resulted in several constraints. First, because we are working from another team’s digitization efforts which produced images of varying quality, we have had to select cases more for image quality than for interest or hand. This has meant that while (for example) we might have wished to draw more from coroners’ rolls because they offer a wide variety of hands, the quality of the images on the AALT made them poor training data. Second, the AALT has only digitized crown cases, which means that cases from manorial and other local courts have been excluded. However, because both Latin and the documentary script used for its records were standard for almost all legal records of the period, whether local or crown, the AALT examples nevertheless end up providing a good representation for the wider documentary practice of the period. We thus believe that this tool has immediate application on manorial and local records yet to be digitized.

\subsection{Test set}
We chose a test set of twenty cases comprised of about half “standard” cases: ones where the handwriting and abbreviations were of average difficulty and where the subject was not very unusual. We imagine such cases would make up most of what researchers would be interested in (for an example, see KB27 263m21). The other half was made up of “difficult” and “unusual” cases, ones that covered less common subject matter (like JUST1 633m52, a complex case of debt and book theft) or had especially difficult hands (like JUST1 230m2). Although we imagine researchers will use the tool to handle more cases like the former than cases like the latter, we included the “unusual” half of the test set to test the model’s effectiveness on edge cases that are difficult even for human paleographers. The test set that we chose does not overlap with the training set, but there might be a few cases in which the same scribe was responsible for both a test case and a training case, because many of the scribes (in particularly for the King's Bench) wrote many rolls. Among the JUST1 and CP examples this is highly unlikely.

\section{The pipeline: design and training}
Our pipeline is available as a web portal at \url{https://glyphmachina.com}.  The user can upload an image, crop it to the relevant case, have a neural network identify the bottoms of all lines of text (the ``baselines''), and have another neural network transcribe the text.  All of the above takes mere seconds.  The user can then send it to Gemini Pro 3 to correct transcription errors, which takes a few minutes.  The user can also send the resulting Latin transcription to Gemini for translation into modern English.  In this section, we briefly describe the main steps of the pipeline, with a focus on the transcription step.

\subsection{Segmentation}
For our purposes, segmentation only involves extracting images of individual lines of text from a manuscript image.  The vast majority of medieval English legal documents have a simple layout, and our web portal allows users to crop out only the block of text they are interested in.  For these reasons, we adopt a simple segmentation algorithm which works well in practice.

We isolate line images by training a baseline segmentation model using the OCR engine kraken.\footnote{\url{https://github.com/mittagessen/kraken}}  We use the default R-Blla architecture \cite{kiessling_2020}, except that all images are resized to a width of 1800 pixels instead of a height of 1800 pixels.  The R-Blla architecture consists of 5 convolutional layers connected to two LSTM layers, one for the x dimension and one for the y dimension, followed by a 1x1 convolution bottleneck and another two LSTMs (for the y and x directions).  The output consists of 3 heat maps: one indicates the beginnings of baselines, one indicates the baselines themselves, and the third indicates the ends.  During inference, kraken finds the baselines from these heat maps and draws polygons around the text lines.  We optimized this portion of the pipeline, reducing its runtime from 52 s to 2.6 s on an Intel Core i9 13900k CPU and a RTX 4090 GPU.  Most of the reduced runtime came from removing the polygon-drawing algorithm; most of the rest came from offloading CPU-intensive image processing tasks to the GPU using cuCIM.\footnote{\url{https://github.com/rapidsai/cucim}}

After identifying the baselines, we extract line images by defining regions less than $0.73H$ pixels above and less than $0.23H$ pixels below the baseline as part of the text line, where H is the median spacing between baselines.  We then rectify the line images by shifting the pixels in each column up or down using linear interpolation so that the baseline in the rectified image is horizontal.  Finally, we convert the image to grayscale, normalize it by subtracting the pixel value median and dividing by the difference between the 90th and 10th percentiles, and invert the result so that text is bright while the background is near 0.

\begin{figure*}[ht]
  \centering \subfigure {\includegraphics
    [width=\textwidth]{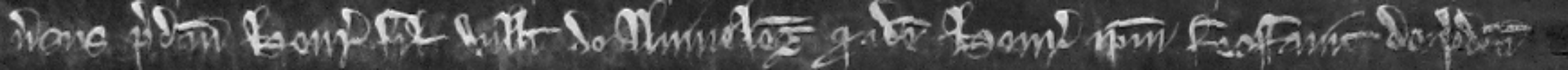}}\qquad\subfigure {\includegraphics
    [width=\textwidth]{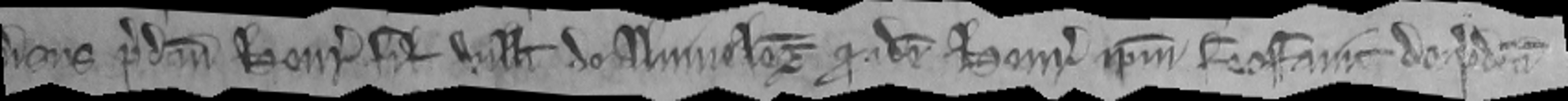}}
    \caption{A line image generated by our algorithm (top) and by kraken (bottom).  We obtain 4\% lower word error rate using our own algorithm. Image from: JUST1 734m1, line 19.}
\label{fig:line_images}
\end{figure*}

Our line image extraction method is far simpler than \textit{kraken}'s sophisticated algorithm, which uses the baseline as a starting point to draw a polygon around the text line, maximizing the text included inside the polygon while minimizing overlap with adjacent lines.  kraken's algorithm can handle lines of very different heights, and includes ascenders and descenders to the maximum extent possible while zeroing out strokes from adjacent lines (see Figure \ref{fig:line_images}).  We expected superior transcription accuracy using kraken's line images, but observed the opposite: the word error rate of the CNN + LSTM model (Table \ref{table:performance_comparison_models}) was 4\% lower using our custom line images.  We suspect this is due to a combination of the strong gradient between the zero filling and the line image, variability the vertical location of the baseline, and inconsistency of the unsubtracted background in kraken's line images.  We encourage more research in this area, because nearly every other publicly available dataset of medieval Latin handwriting (e.g. TRIDIS, CATMuS) uses line images generated by a very similar algorithm to kraken's.

\subsection{Transcription}
We implemented two very different models to transcribe the line images.  The first is a classic CNN + LSTM, with a CTC decoder coupled with an n-gram language model.  The second is the transformer-based TrOCR.

\subsubsection{CNN + LSTM}
This model is based on the architecture in \cite{aguilar_2023}.  Line images are resized to a height of 128 (preserving their aspect ratio) and fed into four convolutional blocks.  Each block consists of a 2D convolution layer, a ReLU activation function, a 2D batch normalization step, and a max pool with a 2x2 kernel size and 2x2 stride.  We replaced the dropout layers in \cite{aguilar_2023} with batch normalization layers because we find that this change greatly improves ease of training while having no perceptible negative impact on the test set character error rate or word error rate.

We feed the output of the four convolutional blocks into three bidirectional LSTM layers with dropout layers sandwiched in between.  Each LSTM layer has a latent dimension size of 512 per direction (twice the size of the LSTMs in \cite{aguilar_2023}), meaning that the final output is of size 1024 x N.  A fully connected layer converts this to ($(N_{\rm char} + 1) \times N$ logits, indicating the likelihood that a given column is occupied by a given character (the $N_{\rm char}$ rows) or marks the beginning of a new character (the remaining row).  The details of the architecture are laid out in Table \ref{table:architecture}.

\begin{table}[ht]
\centering
\caption{The fiducial model architecture (CNN + LSTM). All convolutional blocks include ReLU and BatchNorm2d, and all convolution blocks except the last end with a 2$\times$2 max pool with 2$\times$2 stride. The LSTM layers include Dropout ($p=0.3$). Total parameters: 19M.}
\label{table:architecture}
\setlength{\tabcolsep}{10pt} 
\begin{tabular}{@{} l l l @{}} 
\toprule
Stage & Layer Type & Configuration \\
\midrule
1 & Conv Block & $1 \to 32$ ch, k: $4\times16$, pad: $1\times7$, Max Pool \\
2 & Conv Block & $32 \to 32$ ch, k: $4\times16$, pad: $1\times7$, Max Pool \\
3 & Conv Block & $32 \to 64$ ch, k: $3\times8$, pad: $1\times3$, Max Pool \\
4 & Conv Block & $64 \to 64$ ch, k: $3\times8$, pad: $1\times3$ \\
\addlinespace 
5 & Bi-LSTM ($\times 3$) & Latent size: 512 (per direction), Input: 960 \\
6 & Linear & Input: $1024 \to$ Output: $N_{\rm char} + 1$ \\
\bottomrule
\end{tabular}
\end{table}

The neural network is implemented in PyTorch and trained on the $\sim$4000 line training set using the AdamW optimizer and Connectionist Temporal Classification (CTC) loss function.  We start with a learning rate of $10^{-3}$ and weight decay of $10^{-2}$ but reduce the learning rate by a factor of 3 every time the WER stops decreasing for 10 consecutive epochs, down to a minimum learning rate of $10^{-5}$.  We train for 250 epochs, but the test set CER and WER change negligibly after $\sim$100 epochs.

We use data augmentation during training to improve performance.  During each epoch of training, each line image in the training dataset is randomly jittered in brightness, contrast, saturation, and hue by up to 50\%.  We then apply a random affine transformation: each image is rotated by up to $\pm0.7^\circ$, translated by up to $\pm0.01W$ horizontally and $\pm0.02H$ vertically, and scaled by up to 2\%.  In the final step, we either do nothing (1/4 probability), sharpen the image by a factor of 2 (1/4 probability), or apply a Gaussian blur with a randomly chosen standard deviation of 1--6 pixels.

For inference, we find that greedy CTC decoding performs well and beam search has a negligible effect on the results.  However, coupling the beam search with a language model lowers the word error rate by 3\%.  We used pyctcdecode\footnote{\url{https://github.com/kensho-technologies/pyctcdecode}} to perform the beam search decoding with the help of a KenLM\footnote{\url{https://kheafield.com/code/kenlm/}} n-gram model.  We adopted n=2 and trained the KenLM model solely on the text we used to train the neural network. 
 Training the n-gram model on the massive \textit{Corpus Corporum}, a Latin library containing 215 million words, did not improve results; in fact, it increased the WER by 1\%.

\subsubsection{TrOCR} TrOCR \cite{li_2023} is an entirely transformer-based model which comes in three sizes: small (62M parameters), base (334M parameters), and large (558M parameters).  Training even the small model from scratch is infeasible with our limited training set.  \cite{li_2023} pretrained TrOCR with 684M textlines from PDF files on the Internet, followed by 17.9M synthetic textlines in 5427 handwritten fonts, with the text taken from Wikipedia.  

Because our training data is far more limited, we finetune the TrOCR model of \cite{aguilar_2025}, which in turn was finetuned from the TrOCR\textsubscript{LARGE} model using 178,000 lines of medieval and early modern handwriting.  We use an initial learning rate of $5 \times 10^{-5}$ which linearly decays to zero and a batch size of 8, and train for 30 epochs.  We tried training for 250 epochs and found that both CER and WER start increasing well before the end of training, indicating overfitting.  Data augmentation is performed on the training data using the same procedure as for the CNN + LSTM architecture.

\subsection{Correction by Gemini}
After obtaining transcriptions of all textlines in the image from our neural network, we call the Gemini API and ask Gemini Pro 3 to correct obvious errors.  We do not provide Gemini the image to be transcribed, only the preliminary transcription.  We provide the following system prompt:

\textit{I ran a handwriting recognition neural network on a medieval English legal case, written in Latin.  Please try to correct the mistakes.  The neural network doesn't typically add or remove entire words, so try to keep a similar word count, if possible.  Also, preserve the line breaks at all costs.  If you see a very short line, don't delete it or merge it with the previous or next line, because it's likely an interlinear addition.  Keep in mind that this is medieval, not classical, Latin.  Please don't add punctuation unless necessary, because these documents usually don't have much punctuation.  Do not output anything other than the corrected transcription--no explanations, comments, or alternatives.}

\section{Results}
In this section, we use the character error rate (CER) and word error rate (WER) to evaluate our models.  We also include qualitative insights from expert paleographers, which show that CER/WER is not the whole story.

\subsection{Quantitative evaluation}

\begin{table}[ht]
\centering
\caption{Performance comparison of different models}
\setlength{\tabcolsep}{8pt}
\begin{tabular}{@{} l c c @{}}
\toprule
Model & CER & WER\\
\midrule
CATMuS & 47 & 81\\
TRIDIS (CNN+LSTM) & 29 & 60\\
TRIDIS (TrOCR) & 38 & 64\\
\midrule
CNN + LSTM & 6.5 & 20.6\\
CNN + LSTM + KenLM & 6.0 & 17.8\\
\textbf{CNN + LSTM + KenLM + Gemini} & \textbf{3.8} & \textbf{11.9}\\
TrOCR & 6.9 & 16.3\\
TrOCR + Gemini & 5.9 & 14.9\\
\bottomrule
\end{tabular}
\label{table:performance_comparison_models}
\end{table}

No off-the-shelf model is capable of producing a useful transcription of our dataset, but our models perform well and offer significant time savings to paleographers. 
 This is evident in Table \ref{table:performance_comparison_models}, which shows the character error rate and word error rate achieved by both our own models (below the horizontal line) and by recently published models trained on medieval handwriting (above the horizontal line).    The standard CNN + LSTM architecture with CTC loss results in an acceptably low word error rate of 21\%.  Adding a simple bigram language model reduces the WER to 18\%, while correction by Gemini 3 Pro dramatically improves the transcription, reducing WER to only 12\%.  The transformer-based TrOCR architecture outperforms CNN + LSTM + KenLM in WER, but not in CER.  After correction by Gemini, the CNN + LSTM + KenLM transcriptions significantly outperform the TrOCR transcriptions in both CER and WER.  This suggests that the classical architecture's errors, even if slightly more numerous, are easier to recognize and correct than TrOCR's errors--a hypothesis confirmed by human examination of the transcriptions (see next subsection).

\begin{table}[ht]
\centering
\caption{Word error rates achieved by the fiducial model (CNN + LSTM + KenLM) on each individual case.  We also compare to the WERs achieved by Gemini alone (see Subsection \ref{subsec:gemini}).}
\label{table:performance_comparison}
\setlength{\tabcolsep}{8pt} 
\begin{tabular}{@{} l c c c l @{}}
\toprule
Case & Type & No Gemini & With Gemini & Gemini only \\
\midrule
CP40-236m280 & Standard   & 15 & 7  & 32 \\
CP40-58m14b  & Difficult   & 25 & 10 & Hallucinated lines \\
CP40-648m81a  & Standard  & 9  & 4  & No output \\
CP40-649m116a  & Difficult & 17 & 9  & 20 \\
JUST1-230m2  & Difficult   & 33 & 21 & 58 \\
JUST1-235m13  & Difficult  & 27 & 20 & 45 \\
JUST1-633m52  & Difficult  & 12 & 8  & No output \\
JUST1-633m5a  & Difficult  & 9  & 8  & 27 \\
JUST1-633m60d & Difficult  & 15 & 9  & Hallucinated lines \\
JUST1-734m17da & Difficult & 28 & 17 & Hallucinated lines \\
JUST1-734m24d & Difficult  & 14 & 10 & 28 \\
JUST1-734m4 & Difficult    & 16 & 9  & Hallucinated lines \\
KB27-263m21   & Standard  & 24 & 21 & 54 \\
KB27-645m22a  & Standard  & 37 & 16 & Hallucinated lines \\
KB27-342Rm22  & Standard  & 14 & 12 & Hallucinated lines \\
KB27-345Rm87a & Standard  & 13 & 10 & Hallucinated lines \\
KB27-345Rm87b  & Standard & 9  & 7  & 88 \\
KB27-359Rm8d  & Standard  & 12 & 9  & Hallucinated lines \\
KB27-344Rm14  & Standard  & 14 & 19 & Hallucinated lines \\
KB27-352Rm13  & Standard  & 14 & 12 & 94 \\
\bottomrule
\end{tabular}
\end{table}


Table \ref{table:performance_comparison} shows the WER achieved by the fiducial model on each case within the test set, both with and without Gemini correction.  Gemini almost always improves the transcription, but curiously, there is one case where it increases the error rate (KB27-344Rm14).  As expected, some cases are easier than others for the neural network to transcribe.  However, all transcriptions are accurate enough to be usable, and the error rate is not dramatically different across the three types of manuscript (Common Pleas, Justices in Assize, King's Bench).  They are also not dramatically different between the cases we selected to be ``standard'' and those we selected to be ``difficult'': the former has a WER of 16.1\% prior to Gemini correction, compared to 19.6\% for the latter.  These rates drop to the nearly identical 11.7\% and 12.1\% after Gemini correction.

\subsection{Qualitative evaluation}
Our team includes three medievalists trained in paleography: EW, CW, and SS. To supplement our quantitative evaluation, we present their subjective evaluation of the quality and usefulness of the transcriptions.

\begin{figure}[ht]
    \centering
    \includegraphics[width=1\linewidth]{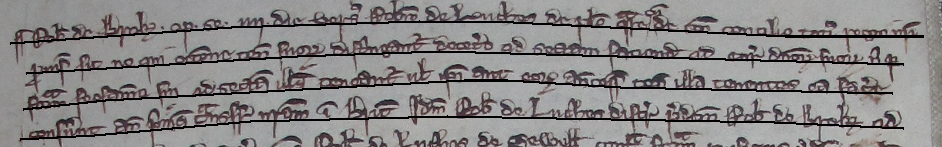}
    \caption{JUST1 701m6}
    \label{fig:evaluation image}
\end{figure}

Some aspects of usefulness to the user are more difficult to characterize quantitatively, and these demonstrate that though the Gemini-corrected model has a lower error rate, it might also mislead researchers where the model will not. To illustrate this, we chose an example of a case from a roll the model had never seen before, from about fifty years before our dataset's range (see Figure \ref{fig:evaluation image}). We chose this example to simulate how a user might stretch the model's capabilities; the case is outside our time period, the handwriting is not very clear, and it has both idiosyncratic abbreviations and smudges. This difficulty was reflected in the accuracy of the transcription the model provided—85\% (in 161 words, the model made 24 errors)—and of the Gemini corrections—94\% (10 errors). 

For the purposes of the user, there are roughly three types of errors, in ascending order of severity: 

\begin{enumerate}
\item A word that is misspelled but easily comprehensible (highlighted in\colorbox{green}{green})

\item An incomprehensible word (highlighted in\colorbox{yellow}{yellow})

\item An incorrect word that is semantically appropriate (underlined and highlighted in\colorbox{red}{\underline{red}})
\end{enumerate}
The first is easily corrected, and the second cues the user to return to the manuscript. But the third is dangerous to understanding, because it does not signal a possible error and might escape notice. 

In the example below, we have taken the first four lines of the test case and highlighted the errors according to their type for both the model's transcription and the Gemini-corrected version (though we have highlighted the entire word, often the mistake is a single letter, as with communi/commune). We then selected a few of the errors to better illustrate the recognition characteristics in a table.

\subsubsection{Model:}
 
\begin{enumerate}
\item Robertus de Brok optulit se iiij die versus\colorbox{green}{Roberto}de Luches de placito quare de \colorbox{yellow}{tonmunti}consilio\colorbox{red}{\underline{toti}}regni nostri 

\item \colorbox{yellow}{peris}sit ne qui occasione\colorbox{green}{tenementa}suorum distringantur decetero ad\colorbox{yellow}{seatm} faciendam ad Curiam dominorum suorum nisi per 

\item formam feoffamenti sui ad sectam illam\colorbox{green}{tencantur}vel ipsi aut eorum\colorbox{green}{antetesie} tenementa illa tenentes ea facere 

\item \colorbox{green}{consuerunt}ante primam transfretacionem nostram et\colorbox{yellow}{bricis}Idem Robertus de Luches distringet predictum Robertum de Brok ad 

\end{enumerate}

\subsubsection{Gemini-corrected:}
\begin{enumerate}
\item	Robertus de Brok optulit se iiij die versus Robertum de Luches de placito quare\colorbox{red}{\underline{cum}}de\colorbox{yellow}{communi}consilio\colorbox{red}{\underline{[]}}regni nostri 

\item	provisum sit ne qui occasione tenementorum suorum distringantur decetero ad sectam faciendam ad Curiam dominorum suorum nisi per 

\item	formam feoffamenti sui ad sectam illam teneantur vel ipsi aut eorum antecessores tenementa illa tenentes\colorbox{red}{\underline{eam}}facere 

\item	consueverint ante primam transfretacionem nostram in Britanniam idem Robertus de Luches distringit predictum Robertum de Brok ad 
\end{enumerate}

\begin{table}
\centering
\caption{Selected errors with Ground Truth, model, and Gemini outputs.}
    \begin{tabular}{c|ccccc}
         &\includegraphics[height=.4in]{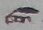}& \includegraphics[height=.4in]{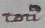}&\includegraphics[height=.4in]{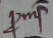}&\includegraphics[height=.4in]{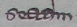}& \includegraphics[height=.4in]{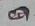}\\
         \hline
         Ground Truth &commune&totius&provisum&sectam&ea\\
         Model &tonmunti&toti&peris&seatm&ea \\ 
 Gemini &communi&[]&provisum&sectam&eam\\
    \end{tabular}
\end{table}

Of the ten errors the model made in these lines, nine were of the first two types (fixable by the user), and only one was of the misleading third type. The Gemini-corrected version made only four errors, but three were misleading, including a total elision (totius) and the introduction of an error where the model made none and where the manuscript is especially clear (ea). For this reason, we recommend that users do not use the Gemini-corrected version without reference to the model's version.  Interestingly, the transformer-based OCR model that we trained (TrOCR) suffers from similar tendencies: it would often guess a grammatically correct Latin word that is not relevant to the case at hand, thus achieving similar word accuracy as our fiducial model, but worse character accuracy.
 
Tests on legal rolls other than JUST1, CP, and KB suggest that the model produces comprehensible transcriptions, but may occasionally be less accurate. It does very well on JUST2 and JUST3 rolls, as well as other KB and CP rolls that were not included in the dataset. It has a lower accuracy on the Exchequer of Pleas rolls, or E rolls, which recorded debts owed to the crown and their payment, possibly due to differences of vocabulary and the many numbers involved. 
   
Testing has shown that the model transcribes texts written earlier than those in our dataset (like Figure 4) with a higher degree of accuracy than texts written later than our dataset. This is likely because the hands and many of the legal forms were standardized before the period of our data set, but after our data set, growing literacy rates, the advent of the printing press, and the prevalence of the vernacular (English and French) in official life increased the variation in the hands and introduced more of the vernacular. 

Currently, the model is trained on Latin only and will not provide accurate transcriptions of documents in vernacular languages. On occasion, however, English words, often job titles such as "skynner" or "yeoman", will appear in predominantly Latin legal enrollments. In such cases, the model will usually transcribe these words correctly, especially if the word is written in full. 

\subsection{Can Gemini 3 Pro transcribe cases by itself?}
\label{subsec:gemini}
When we began the current project in October 2023, no LLMs that we tested were at all capable of transcribing medieval English legal documents.  As of December 2025, GPT 5.1 continues to be unable to transcribe our dataset.  It refuses to even make an attempt for longer cases, and when prompted to attempt just the first few lines, it hallucinates a transcription with no relationship to the manuscript. Claude Sonnet 4.5 likewise refuses to attempt a transcription, while Grok 4.1 hallucinates a transcription.

Gemini Pro 3, however, performs remarkably well.  We have noticed that for some cases, it produces a remarkably good transcription, even as it fails catastrophically on others. In keeping with its inconsistency across cases, Gemini Pro 3's accuracy also alters across iterations of attempts on the same case.  To quantify its performance, we used the Gemini API to request a transcription for each case in the test set, with the following prompt:

\textit{Attached is a medieval English legal case, written in Latin. Please transcribe it line by line in vertical order, and don't output anything other than the transcription. Interlinear additions count as independent lines. Transparently expand all abbreviations and don't put the expansion in square brackets. Don't add punctuation that isn't in the manuscript. When you're not sure, take your best guess instead of putting [?] or indicating multiple possible alternatives.}

Table \ref{table:performance_comparison} shows the resulting WER for each case.  For 3 cases, Gemini produces an acceptable transcription (WER < 30\%), although it does not outperform our model.  For 2 cases, Gemini returns no output.  For 9 cases, Gemini returns the wrong number of text lines.  We examined the output for these 9 cases and found that the mismatch is not the result of minor mistakes, but of Gemini hallucinating text not found anywhere in the manuscript.

We conclude that frontier LLMs are not yet capable of consistently producing reliable transcriptions of medieval English legal cases.  However, Gemini is making rapid progress--Gemini Pro 2.5 was completely unable to produce any reasonable transcriptions, while Gemini Pro 3 sometimes produces transcriptions with a word error rate of only 20\%.  It is not impossible that the next version of Gemini will excel at this task, at least for some scrolls.

\section{Conclusion}
We release a dataset containing $\sim4000$ lines of medieval legal Latin.  This dataset is a sample from the millions of manuscript pages that comprise the English legal tradition, which the legal systems of much of the English-speaking world is based on.  The formulaic nature of these documents, the specialized vocabulary and abbreviations, and the broad cross section of everyday life that these legal cases address make our dataset unique in the literature.  These distinctive features pose both problems and opportunities for machine learning models.

We find that a CNN + LSTM + CTC model trained on this dataset achieves excellent accuracy, especially when coupled with a bigram model.  The model is sufficient to transcribe the majority of the medieval English legal corpus to a high degree of comprehensibility, and most of the errors the model does make do not impact understanding.  Gemini can correct many of the remaining errors, but at the cost of sometimes making detectable errors undetectable.  We release this model to the public in the form of an easy-to-use web portal, vastly increasing the accessibility of the medieval English legal tradition to students and scholars.

\section{Data and code availability}
All the code and data required to reproduce these results are publicly available on \href{https://github.com/ideasrule/glyph_machina_public}{GitHub}.  The data are also hosted on \href{https://huggingface.co/datasets/mzzhang2014/glyph_machina}{Hugging Face}.

\begin{credits}
\subsubsection{\ackname} CW thanks the Ames Foundation for funding her work on this project.

\subsubsection{\discintname}
The authors have no competing interests to declare that are
relevant to the content of this article.
\end{credits}

%
%
%
\bibliographystyle{splncs04}
\bibliography{main}

@article{kim_1997,
author = {Kim, Gyeonghwan and Govindaraju, Venu},
title = {A Lexicon Driven Approach to Handwritten Word Recognition for Real-Time Applications},
year = {1997},
issue_date = {April 1997},
publisher = {IEEE Computer Society},
address = {USA},
volume = {19},
number = {4},
issn = {0162-8828},
url = {https://doi.org/10.1109/34.588017},
doi = {10.1109/34.588017},
journal = {IEEE Trans. Pattern Anal. Mach. Intell.},
month = apr,
pages = {366–379},
numpages = {14}
}

@article{shi_2017,
  author       = {Baoguang Shi and
                  Xiang Bai and
                  Cong Yao},
  title        = {An End-to-End Trainable Neural Network for Image-Based Sequence Recognition
                  and Its Application to Scene Text Recognition},
  journal      = {{IEEE} Trans. Pattern Anal. Mach. Intell.},
  volume       = {39},
  number       = {11},
  pages        = {2298--2304},
  year         = {2017},
  url          = {https://doi.org/10.1109/TPAMI.2016.2646371},
  doi          = {10.1109/TPAMI.2016.2646371},
  bibsource    = {dblp computer science bibliography, https://dblp.org}
}

@ARTICLE{garrido-munoz_2025,
       author = {{Garrido-Munoz}, Carlos and {Rios-Vila}, Antonio and {Calvo-Zaragoza}, Jorge},
        title = "{Handwritten Text Recognition: A Survey}",
      journal = {arXiv e-prints},
         year = 2025,
        month = feb,
          eid = {arXiv:2502.08417},
        pages = {arXiv:2502.08417},
          doi = {10.48550/arXiv.2502.08417},
archivePrefix = {arXiv},
       eprint = {2502.08417},
 primaryClass = {cs.CV},
       adsurl = {https://ui.adsabs.harvard.edu/abs/2025arXiv250208417G}
}

@inproceedings{vaswani_2017,
  author       = {Ashish Vaswani and
                  Noam Shazeer and
                  Niki Parmar and
                  Jakob Uszkoreit and
                  Llion Jones and
                  Aidan N. Gomez and
                  Lukasz Kaiser and
                  Illia Polosukhin},
  editor       = {Isabelle Guyon and
                  Ulrike von Luxburg and
                  Samy Bengio and
                  Hanna M. Wallach and
                  Rob Fergus and
                  S. V. N. Vishwanathan and
                  Roman Garnett},
  title        = {Attention is All you Need},
  booktitle    = {Advances in Neural Information Processing Systems 30: Annual Conference
                  on Neural Information Processing Systems 2017, December 4-9, 2017,
                  Long Beach, CA, {USA}},
  pages        = {5998--6008},
  year         = {2017},
  url          = {https://proceedings.neurips.cc/paper/2017/hash/3f5ee243547dee91fbd053c1c4a845aa-Abstract.html},
  bibsource    = {dblp computer science bibliography, https://dblp.org}
}

@article{kang_2022,
  author       = {Lei Kang and
                  Pau Riba and
                  Mar{\c{c}}al Rusi{\~{n}}ol and
                  Alicia Forn{\'{e}}s and
                  Mauricio Villegas},
  title        = {Pay attention to what you read: Non-recurrent handwritten text-Line
                  recognition},
  journal      = {Pattern Recognit.},
  volume       = {129},
  pages        = {108766},
  year         = {2022},
  url          = {https://doi.org/10.1016/j.patcog.2022.108766},
  doi          = {10.1016/J.PATCOG.2022.108766},
  bibsource    = {dblp computer science bibliography, https://dblp.org}
}

@inproceedings{li_2023,
  author       = {Minghao Li and
                  Tengchao Lv and
                  Jingye Chen and
                  Lei Cui and
                  Yijuan Lu and
                  Dinei A. F. Flor{\^{e}}ncio and
                  Cha Zhang and
                  Zhoujun Li and
                  Furu Wei},
  editor       = {Brian Williams and
                  Yiling Chen and
                  Jennifer Neville},
  title        = {TrOCR: Transformer-Based Optical Character Recognition with Pre-trained
                  Models},
  booktitle    = {Thirty-Seventh {AAAI} Conference on Artificial Intelligence, {AAAI}
                  2023, Thirty-Fifth Conference on Innovative Applications of Artificial
                  Intelligence, {IAAI} 2023, Thirteenth Symposium on Educational Advances
                  in Artificial Intelligence, {EAAI} 2023, Washington, DC, USA, February
                  7-14, 2023},
  pages        = {13094--13102},
  publisher    = {{AAAI} Press},
  year         = {2023},
  url          = {https://doi.org/10.1609/aaai.v37i11.26538},
  doi          = {10.1609/AAAI.V37I11.26538},
  bibsource    = {dblp computer science bibliography, https://dblp.org}
}

@article{aguilar_2023,
    title      = {Handwritten Text Recognition for Documentary Medieval Manuscripts},
    author     = {Sergio Torres Aguilar and Vincent Jolivet},
    url        = {https://jdmdh.episciences.org/10484},
    doi        = {10.46298/jdmdh.10484},
    journal    = {Journal of Data Mining \& Digital Humanities},
    issn       = {2416-5999},
    volume     = {Historical Documents and automatic text recognition},
    eid        = 6,
    year       = {2023},
    month      = {Dec},
    language   = {English},
}

@article{aguilar_2025,
  author       = {Sergio Torres Aguilar},
  title        = {{TRIDIS:} {A} Comprehensive Medieval and Early Modern Corpus for {HTR}
                  and {NER}},
  journal      = {CoRR},
  volume       = {abs/2503.22714},
  year         = {2025},
  url          = {https://doi.org/10.48550/arXiv.2503.22714},
  doi          = {10.48550/ARXIV.2503.22714},
  eprinttype    = {arXiv},
  eprint       = {2503.22714},
  bibsource    = {dblp computer science bibliography, https://dblp.org}
}

@INPROCEEDINGS{kiessling_2020,
  author={Kiessling, Benjamin},
  booktitle={2020 17th International Conference on Frontiers in Handwriting Recognition (ICFHR)}, 
  title={A Modular Region and Text Line Layout Analysis System}, 
  year={2020},
  volume={},
  number={},
  pages={313-318},
  doi={10.1109/ICFHR2020.2020.00064}}

@inproceedings{fujitake_2024,
  author       = {Masato Fujitake},
  title        = {DTrOCR: Decoder-only Transformer for Optical Character Recognition},
  booktitle    = {{IEEE/CVF} Winter Conference on Applications of Computer Vision, {WACV}
                  2024, Waikoloa, HI, USA, January 3-8, 2024},
  pages        = {8010--8020},
  publisher    = {{IEEE}},
  year         = {2024},
  url          = {https://doi.org/10.1109/WACV57701.2024.00784},
  doi          = {10.1109/WACV57701.2024.00784},
  bibsource    = {dblp computer science bibliography, https://dblp.org}
}

@article{diaz_2021,
  author       = {Daniel Hernandez Diaz and
                  Siyang Qin and
                  R. Reeve Ingle and
                  Yasuhisa Fujii and
                  Alessandro Bissacco},
  title        = {Rethinking Text Line Recognition Models},
  journal      = {CoRR},
  volume       = {abs/2104.07787},
  year         = {2021},
  url          = {https://arxiv.org/abs/2104.07787},
  eprinttype    = {arXiv},
  eprint       = {2104.07787},
  bibsource    = {dblp computer science bibliography, https://dblp.org}
}

@InProceedings{clerice_2024,
author="Cl{\'e}rice, Thibault
and Pinche, Ariane
and Vlachou-Efstathiou, Malamatenia
and Chagu{\'e}, Alix
and Camps, Jean-Baptiste
and Levenson, Matthias Gille
and Brisville-Fertin, Olivier
and Boschetti, Federico
and Fischer, Franz
and Gervers, Michael
and Boutreux, Agn{\`e}s
and Manton, Avery
and Gabay, Simon
and O'Connor, Patricia
and Haverals, Wouter
and Kestemont, Mike
and Vandyck, Caroline
and Kiessling, Benjamin",
editor="Barney Smith, Elisa H.
and Liwicki, Marcus
and Peng, Liangrui",
title="CATMuS Medieval: A Multilingual Large-Scale Cross-Century Dataset in Latin Script for Handwritten Text Recognition and Beyond",
booktitle="Document Analysis and Recognition - ICDAR 2024",
year="2024",
publisher="Springer Nature Switzerland",
address="Cham",
pages="174--194",
isbn="978-3-031-70543-4"
}

@INPROCEEDINGS{kahle_2017,
  author={Kahle, Philip and Colutto, Sebastian and Hackl, Günter and Mühlberger, Günter},
  booktitle={2017 14th IAPR International Conference on Document Analysis and Recognition (ICDAR)}, 
  title={Transkribus - A Service Platform for Transcription, Recognition and Retrieval of Historical Documents}, 
  year={2017},
  volume={04},
  number={},
  pages={19-24},
  doi={10.1109/ICDAR.2017.307}
}

@misc{puigcerver_2018,
  author = {Joan Puigcerver and Carlos Mocholí},
  title = {PyLaia},
  year = {2018},
  publisher = {GitHub},
  journal = {GitHub repository},
  howpublished = {\url{https://github.com/jpuigcerver/PyLaia}},
  commit = {commit SHA}
}

@article{nockels_2022,
  author    = {Nockels, Joe and Gooding, Paul and Ames, Sarah and Terras, Melissa},
  title     = {Understanding the application of handwritten text recognition technology in heritage contexts: a systematic review of Transkribus in published research},
  journal   = {Archival Science},
  year      = {2022},
  volume    = {22},
  number    = {3},
  pages     = {367--392},
  doi       = {10.1007/s10502-022-09397-0},
  url       = {https://doi.org/10.1007/s10502-022-09397-0},
  issn      = {1573-7500}
}

@InProceedings{coll-ardanuy_2026,
author="Coll Ardanuy, Mariona
and Berganzo-Besga, Iban
and Sarobe, Ramon
and Cuadrada, Coral",
editor="Yin, Xu-Cheng
and Karatzas, Dimosthenis
and Lopresti, Daniel",
title="Evaluating Handwritten Text Recognition in Medieval Notarial Manuscripts: A New Dataset and Comprehensive Analysis",
booktitle="Document Analysis and Recognition -- ICDAR 2025",
year="2026",
publisher="Springer Nature Switzerland",
address="Cham",
pages="340--357",
isbn="978-3-032-04624-6"
}

@inproceedings{pugh_1983,
    author = "Pugh, Ralph B.",
    title = "The Duration of Criminal Trials in Medieval England",
    editor = "Manchester, A.H. 
and Ives, Eric William",
    booktitle = "Law, Litigants and the Legal Profession",
    publisher = "Royal Historical Society",
     pages = "104--115",
    year = "1983"
}

\end{document}